\documentclass[sigconf]{acmart}

\usepackage{multirow}
\usepackage[linesnumbered,ruled,vlined]{algorithm2e}
\usepackage{booktabs}
\usepackage{tabularx}
\usepackage{makecell}
\usepackage{caption}
\usepackage{adjustbox}
\usepackage{xspace}
\usepackage{enumitem}
\usepackage[most]{tcolorbox}    
\newcommand{\hide}[1]{}

\newcommand{\OursMethod}{\textbf{TESLA}\xspace}
\newcommand{\OursDataset}{\textbf{CASCADE}\xspace}

\newcommand{\acronym}[1]{\underline{\textbf{#1}}}

\newcommand{\feature}{\mathbf{x}}
\newcommand{\conLabel}{y}
\newcommand{\refLabel}{z}

\newcommand{\loss}{\mathcal{L}}

\newcommand{\click}{c}
\newcommand{\conversion}{v}
\newcommand{\refund}{r}
\newcommand{\net}{n}
\newcommand{\observe}{o}

\newcommand{\share}{s}
\newcommand{\pro}{p}

\newcommand{\window}{W}

\newcommand{\weight}{w}
\newcommand{\TIME}{t}

\newcommand{\delay}{h}
\newcommand{\interval}{e}

\newcommand{\realPro}{p}
\newcommand{\obsPro}{q}

\newcommand{\posSet}{\mathcal{P}}
\newcommand{\negSet}{\mathcal{N}}

\newcommand{\encodings}{ \mathrm{Enc}} 
\newcommand{\embeddings}{ \mathrm{Emb}}

\copyrightyear{2026}
\acmYear{2026}
\setcopyright{cc}
\setcctype{by}
\acmConference[WWW '26]{Proceedings of the ACM Web Conference 2026}
{April 13--17, 2026}{Dubai, United Arab Emirates}
\acmBooktitle{Proceedings of the ACM Web Conference 2026 (WWW '26),
April 13--17, 2026, Dubai, United Arab Emirates}
\acmDOI{}
\acmISBN{}
\begin{document}

\title{Modeling Cascaded Delay Feedback for Online Net Conversion Rate Prediction: Benchmark, Insights and Solutions}
\thanks{This paper has been accepted by the ACM Web Conference (WWW) 2026.
This is the camera-ready version.
Please refer to the published version for citation once available.}

\author[]{Mingxuan Luo}
\authornote{The authors contribute equally.}
\orcid{0009-0006-7659-1320}
\email{luomingxuan@stu.xmu.edu.cn}
\affiliation{
  \institution{Xiamen University}
  \city{Xiamen}
  \country{China}
}
\author[]{Guipeng Xv}
\authornotemark[1]
\orcid{0000-0001-5320-5489} 
\email{xuguipeng@stu.xmu.edu.cn}
\affiliation{
  \institution{Xiamen University}
  \city{Xiamen}
  \country{China}
}

\author[]{Sishuo Chen}
\orcid{0009-0007-8845-5817}
\email{chensishuo.css@alibaba-inc.com}
\affiliation{
  \institution{Taobao \& Tmall Group of Alibaba}
  \city{Beijing}
  \country{China}
  }

\author[]{Xinyu Li}
\orcid{0009-0007-5575-8221}
\email{xinyuli@stu.xmu.edu.cn}
\affiliation{
  \institution{Xiamen University}
  \city{Xiamen}
  \country{China}
}

\author[]{Li Zhang}
\orcid{0009-0005-1311-4116}
\email{zl428934@alibaba-inc.com}
\affiliation{
  \institution{Taobao \& Tmall Group of Alibaba}
  \city{Beijing}
  \country{China}
  }

\author[]{Zhangming Chan}
\orcid{0000-0002-3081-2427}
\email{zhangming.czm@alibaba-inc.com}
\affiliation{
  \institution{Taobao \& Tmall Group of Alibaba}
  \city{Beijing}
  \country{China}
  }

\author[]{Xiang-Rong Sheng}
\orcid{0009-0006-4864-574X}
\email{xiangrong.sxr@alibaba-inc.com}
\affiliation{
  \institution{Taobao \& Tmall Group of Alibaba}
  \city{Beijing}
  \country{China}
  }

\author[]{Han Zhu}
\orcid{0000-0002-9522-5637}
\email{zhuhan.zh@alibaba-inc.com}
\affiliation{
  \institution{Taobao \& Tmall Group of Alibaba}
  \city{Beijing}
  \country{China}
  }

\author[]{Jian Xu}
\orcid{0000-0003-3111-1005}
\email{xiyu.xj@alibaba-inc.com}
\affiliation{
  \institution{Taobao \& Tmall Group of Alibaba}
  \city{Beijing}
  \country{China}
  }

\author[]{Bo Zheng}
\orcid{0000-0002-4037-6315}
\email{bozheng@alibaba-inc.com}
\affiliation{
  \institution{Taobao \& Tmall Group of Alibaba}
  \city{Beijing}
  \country{China}
  }

\author[]{Chen Lin}
\authornote{Corresponding author.}
\email{chenlin@xmu.edu.cn}
\orcid{0000-0002-2275-997X}
\affiliation[]{%
  \institution{Xiamen University}
  \city{Xiamen}
  \country{China}
}


\renewcommand{\shortauthors}{Mingxuan Luo et al.}

\begin{abstract} 
In industrial recommender systems, conversion rate (CVR) is often used for traffic allocation, but fails to fully reflect recommendation effectiveness as it does not account for refund rate (RFR). 
Thus, net conversion rate (NetCVR), the probability that a clicked item is purchased and not refunded, is proposed to better show true user satisfaction and business value.
Unlike CVR, NetCVR prediction involves a more complex \textbf{multi-stage cascaded delay feedback phenomenon}. 
The two cascaded delays $\text{Click} \to \text{Conversion}$ and $\text{Conversion} \to \text{Refund}$ in NetCVR have opposite effects. 
Therefore, traditional CVR methods cannot be directly applied.
At present, the lack of relevant open-source datasets and online continuous training schemes poses a challenge.

\begin{sloppypar}
To address these, we first introduce \acronym{CA}scadal \acronym{S}equences of \acronym{C}onversion \acronym{A}nd \acronym{D}elayed r\acronym{E}fund (\OursDataset), the first large-scale open dataset derived from Taobao app for online continuous NetCVR prediction. 
We further analyze \OursDataset and derive three key insights: (1) NetCVR exhibits clear temporal patterns necessitating online continuous modeling; 
(2) Cascaded modeling CVR and RFR for NetCVR outperforms directly modeling NetCVR; 
and (3) delay time, which correlated with both CVR and RFR, is an important feature for NetCVR prediction. 
Based on these insights, we propose ne\acronym{T} conv\acronym{E}rsion ca\acronym{S}caded mode\acronym{L}ing and debi\acronym{A}sing method (\OursMethod).
This continuous method features a CVR-RFR cascaded architecture, stage-wise debiasing, and a delay-time-aware ranking loss for efficient NetCVR prediction.
Experiments show that \OursMethod outperforms state-of-the-art methods on \OursDataset, achieving an absolute improvement of \textbf{12.41\%} in RI-AUC and \textbf{14.94\%} in RI-PRAUC on NetCVR over the strongest baseline.
We hope this work provides a new direction for online delayed feedback modeling in NetCVR prediction.
Our code and dataset are available at \url{https://github.com/alimama-tech/NetCVR}.
\end{sloppypar}

\end{abstract}

\ccsdesc[500]{Information systems~Online advertising}
\ccsdesc[500]{Applied computing~Electronic commerce}

\keywords{Net Conversion Rate Prediction, Cascaded Delayed Feedback, Continuous Learning}

\maketitle

\section{Introduction}
\label{sec:intro}
\begin{figure}[t]
\centering
\includegraphics[width=\columnwidth]{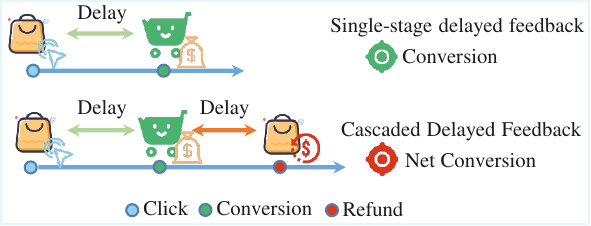}
\caption{Illustration of Net Conversion and Conversion. Net conversion shows multi-stage cascaded delay nature, involving Click $\to$ Conversion and Conversion $\to$ Refund delays. } \label{fig:intro}
\end{figure}

In industrial recommender systems, the post-click conversion rate (CVR) \textemdash defined as the probability that a user makes a purchase after clicking an item \textemdash is commonly predicted and leveraged for traffic allocation~\cite{chapelle2014modeling,yasui2020feedback,chen2022asymptotically,yang2021capturing,gu2021real,wang2023unbiased,liu2023online,liu2024online}.
However, CVR alone may not fully capture the true effectiveness of a recommendation algorithm, as users often request refunds when purchased items fail to meet their expectations.
To better reflect genuine user satisfaction and business value, it is essential to consider not only whether a purchase occurs, but also whether it is retained \textemdash i.e., not followed by a refund.
This motivates the prediction of the net conversion rate (NetCVR)~\cite{zhao2023entire}, which measures the probability that a user completes a purchase and does not subsequently request a refund (as illustrated in Fig.~\ref{fig:intro}).


NetCVR prediction is significantly more challenging than conventional conversion rate (CVR) estimation due to its multi-stage, cascaded delay feedback structure. 
To obtain a definitive training label, one must observe the full user journey: 
$\text{Click} \rightarrow \text{Conversion} \rightarrow \text{Refund}$. 
This process involves two distinct delay intervals: (1) the time between a click and a potential conversion, and (2) the subsequent time between conversion and a potential refund. 

Refunds can occur weeks after purchase, but recommender systems need up-to-date models. 
Existing methods, such as ECAD~\cite{zhao2023entire}, use fully observed labels and adopt batch-based feedback modeling, typically updating the model once per day offline.

However, offline training fails to capture temporal dynamics and lacks the responsiveness required in real-world recommender systems, where models must continuously adapt to evolving user behavior and market conditions~\cite{chapelle2014modeling,ma2018entire,lee2012estimating,fei2025entire,zhang2024adversarial,su2024ddpo,wang2022escm2,wang2025fim}. 
Meanwhile, offline methods cannot be simply extended to online learning scenarios, as they fail to address issues like stream data modeling and debiasing. 
Consequently, online continuous approaches to NetCVR prediction remain largely unexplored, despite their critical importance for practical deployment.



In this work, we present \acronym{CA}scadal \acronym{S}equences of \acronym{C}onversion \acronym{A}nd \acronym{D}elayed r\acronym{E}fund (\OursDataset), \textbf{the first large-scale open dataset dedicated to online continuous net-conversion prediction}. 
We derive \OursDataset using display advertising data collected from the Taobao app, one of the world's largest e-commerce platforms.
It meticulously records the full user interaction sequence, from clicks to conversions and refunds, providing a comprehensive view of user behavior. 
Each event is timestamped, enabling accurate modeling of label evolution under cascaded delayed feedback. 
These features make it an invaluable resource for advancing delayed feedback models in NetCVR prediction.

To gather insights for algorithm development, we perform data analysis and experiments on \OursDataset, leading to three key findings.
(1) NetCVR exhibit \textit{distinct temporal patterns}, underscoring the importance of \textit{online continuous training}.
(2) Cascaded modeling CVR and refund rate (RFR) for NetCVR outperforms direct modeling NetCVR, showing significant potential.
(3) Delay time correlates with both CVR and RFR. Shorter delays are common in high CVR and high RFR cases, suggesting that delay time could be useful for improving NetCVR prediction.

\begin{sloppypar}
Based on the above observations, we present \textbf{the first online continuous method for net conversion prediction}, 
ne\acronym{T} conv\acronym{E}rsion ca\acronym{S}caded mode\acronym{L}ing and debi\acronym{A}sing method (\OursMethod).
To address the time-varying nature of NetCVR, \OursMethod employs online continuous training and constructs a novel data stream for NetCVR prediction.
Since CVR and RFR are highly correlated, \OursMethod employs a shared-bottom architecture to model CVR and RFR as separate tasks with partial shared parameters. 
Additionally, a delay-time-aware ranking loss is employed for real-time training on partial feedback. 
Extension experiments confirm that \OursMethod surpasses state-of-the-art methods on \OursDataset, achieving an absolute improvement of \textbf{12.41\%} in RI-AUC and \textbf{14.94\%} in RI-PRAUC on NetCVR over the strongest baseline.
\end{sloppypar}

We hope this work can offer a new direction for online delayed feedback modeling in NetCVR prediction. We summarize our contributions as follows:
\begin{itemize}[leftmargin=10pt,topsep=2pt]
  \item \textbf{Data Resources}. We introduce the first public benchmark dataset \OursDataset for online continuous NetCVR prediction, offering a reproducible streaming experimental environment. We plan to release our code and benchmark post-publication. 
  
  \item \textbf{Valuable Insights}. Based on data and experimental analysis, we reveal the limitations of current methods in handling cascaded delayed feedback and propose several key insights to guide future research directions.
  
  \item \textbf{Efficient Method}. We evaluate multiple methods on \OursDataset and introduce the first online continuous training framework specifically designed for NetCVR.
  Extensive evaluations show that \OursMethod consistently boosts NetCVR prediction performance.
\end{itemize}

\section{Related Work}
\label{sec:RelatedWork}

\noindent\textbf{Conversion Rate Prediction.}
Click-through conversion rate (CVR) prediction is a core component in recommender systems and online advertising, directly affecting ranking quality and platform revenue~\cite{cheng2016wide,guo2017deepfm,ma2018entire,zhao2023entire,chan2023capturing}. 
Existing efforts have advanced CVR modeling from multiple angles, including model architecture design~\cite{cheng2016wide,guo2017deepfm,ma2018modeling}, multi-task learning with auxiliary signals~\cite{ma2018entire,zhao2023entire,su2024ddpo,wang2022escm2,chen2025see}, and delayed feedback handling~\cite{chapelle2014modeling,gu2021real,wang2023unbiased,liu2023online,liu2024online}. 
While these methods improve estimation accuracy along the click-to-conversion path, they do not account for post-conversion refund behaviors, a critical factor in determining true user satisfaction and net revenue.

\noindent\textbf{Delayed Feedback in Conversion Rate Prediction.}
Methods for addressing delayed feedback fall into three main paradigms:  
(1) \textit{Parametric Survival Modeling}: Early approaches such as DFM~\cite{chapelle2014modeling} and NoDeF~\cite{yoshikawa2018nonparametric} assume parametric delay distributions (e.g., exponential) and jointly model conversion probability and delay time.  
(2) \textit{Label Correction with Importance Sampling}: To correct bias from censored negatives, methods like FNW~\cite{ktena2019addressing} and Defer~\cite{gu2021real} re-weight unconverted samples upon delayed feedback using importance weighting.  
(3) \textit{Hybrid Window-Based Methods}: These balance bias and variance by combining unbiased short-term estimation within an observation window and debiased long-term correction beyond it. Examples include Defuse~\cite{chen2022asymptotically} and DDFM~\cite{dai2023dually}, while DFSN~\cite{liu2023online} and MISS~\cite{liu2024online} further enhance performance via multi-stream modeling across different delay stages.

\noindent\textbf{Net Conversion and Refund-Aware Modeling.}
Recent work has begun to recognize that final user outcomes extend beyond conversion. 
ECAD~\cite{zhao2023entire} introduces net conversion by incorporating refunds, but operates offline with static labels, ignoring the cascaded and asynchronous nature of real-time feedback. 
In contrast, \OursMethod is the first method designed for online continuous NetCVR prediction under multi-stage cascaded delay feedback.

\section{Dataset Construction and Analysis}
\label{sec:Dataset}

\subsection{Dataset Construction}\label{sec:DataConstruct}
\begin{sloppypar}
Although the prediction of net conversion rates (NetCVR) is crucial, one limitation is the absence of open-access datasets.
Existing datasets such as Criteo\footnote{\url{https://labs.criteo.com/2013/12/conversion-logs-dataset/}} and Tencent\footnote{\url{https://algo.qq.com/?lang=en}} provide click and conversion data, but none include post-purchase behavioral signals, such as whether a purchase is retained or refunded, making it impossible to accurately model NetCVR.
To address these limitations, we have developed the \textbf{first open-source dataset for net conversion rate prediction}, named \acronym{CA}scadal \acronym{S}equences of \acronym{C}onversion \acronym{A}nd \acronym{D}elayed r\acronym{E}fund (\OursDataset). 
Detailed construction, feature descriptions, and statistics are provided in Appendix~\ref{apx:data_construction}.
\end{sloppypar}

\begin{figure}[t]
\centering
\includegraphics[width=1\columnwidth]{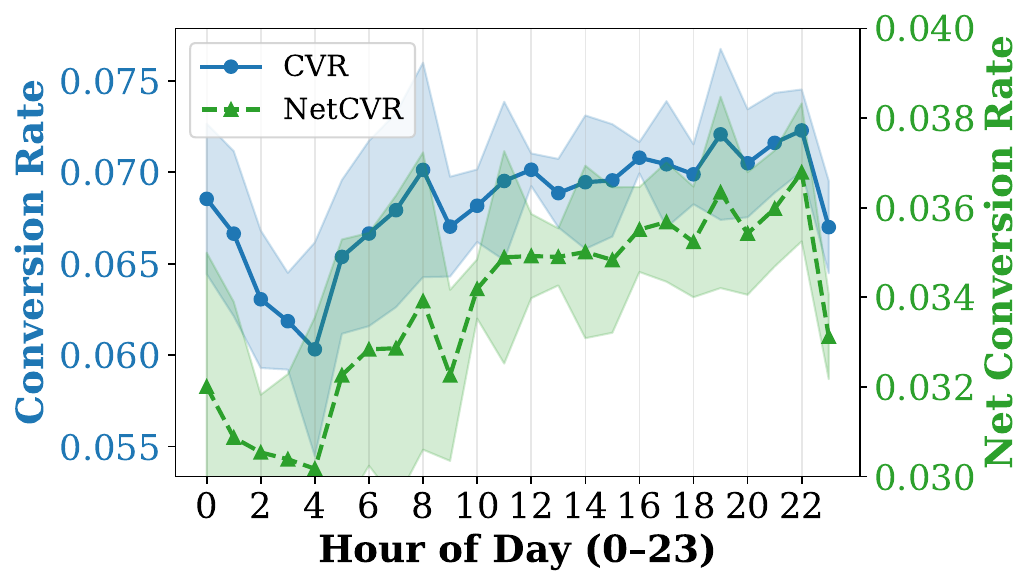} 
%
\caption{Conversion Rate and Net Conversion Rate by Hours. } \label{fig:hourly_cvr_netcvr_std}
\end{figure}
\begin{figure}[t]
\centering
\includegraphics[width=1 \columnwidth]{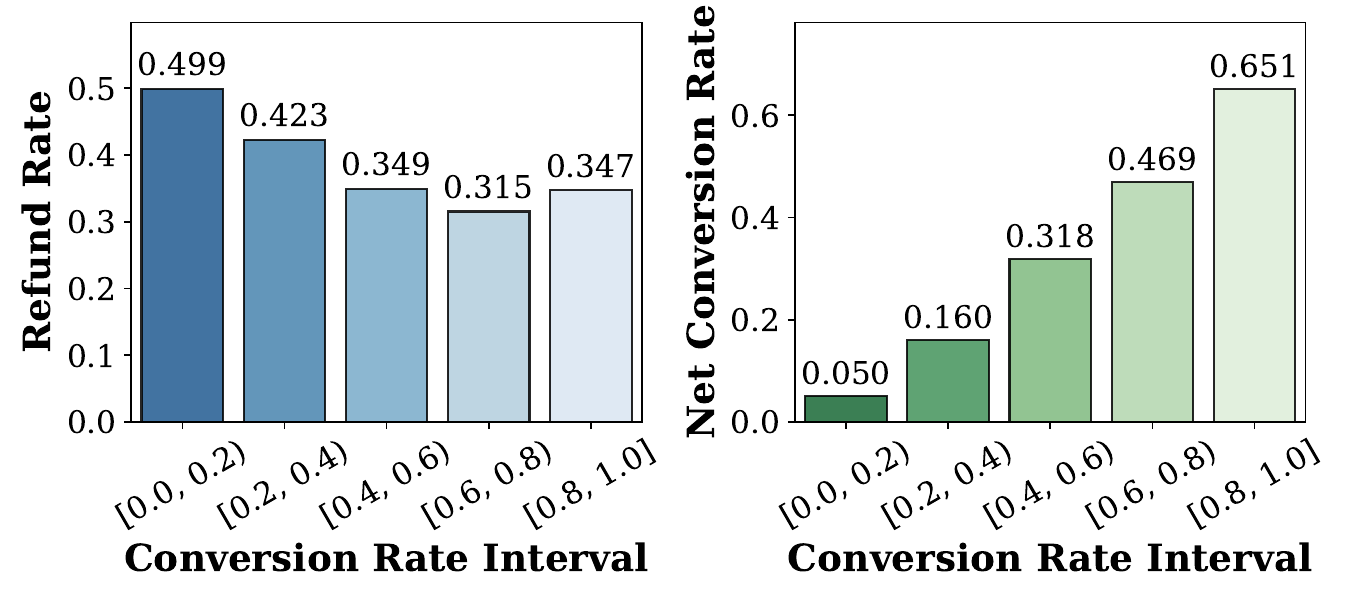}
\caption{Average Refund Rate and Net Conversion Rate Across Conversion Rate Bins.} \label{fig:cvr_rfr_corr}
\end{figure}
%


\subsection{Problem Setup}\label{sec:problemDefine}
In the Conversion Rate (CVR), Refund Rate (RFR), and Net Conversion Rate (NetCVR) prediction task, the model processes inputs as $(\feature, \conLabel, \refLabel)\sim (X, Y, Z)$. 
Here, $\feature$ represents features, and $\conLabel \in \{0, 1\}$, $\refLabel \in \{0, 1\}$ are the conversion and refund labels.
Since net conversion has multi-stage cascaded delay characteristics, we do not collect net conversion labels separately but represent them using conversion and refund labels.
$\conLabel = 1$ and $\refLabel = 1$ represent there exist a conversion and a refund, respectively. 
Net conversion occurs when $\conLabel = 1$ and $\refLabel = 0$. 
Given features $\feature$, the objectives are as follows: 
For CVR, train a parameterized function $f_{\theta}^{\text{CVR}}$ to predict the conversion probability; 
for RFR, train $f_{\theta}^{\text{RFR}}$ to predict the refund probability; 
and for NetCVR, train $f_{\theta}^{\text{NetCVR}}$ to predict the net conversion probability, which can also be represented as $f_{\theta}^{\text{NetCVR}} = f_{\theta}^{\text{CVR}} *(1 - f_{\theta}^{\text{RFR}})$. 
The key notations used in this paper are summarized in Appendix~\ref{apx:notation}.

\subsection{Insights Exploration}\label{sec:DataAnalysis}
Given the lack of prior art in NetCVR prediction, to gain deeper insights, we employ exploratory data analysis in this section. Key insights are listed as follows:  
\begin{tcolorbox}[colback=cyan!5!white, colframe=cyan!45!blue!60, title=\textbf{Takeaways}]
\begin{enumerate}[label=\arabic*., left=0pt]
  \item NetCVR exhibits clear temporal patterns necessitating online continuous learning.
  \item Cascaded modeling CVR and RFR for NetCVR outperforms directly modeling NetCVR. 
  \item Delay time, which is correlated with CVR and RFR, is an important feature for NetCVR prediction.
\end{enumerate}
\end{tcolorbox}

\begin{figure*}[t]
\centering
\includegraphics[width=0.95\textwidth]{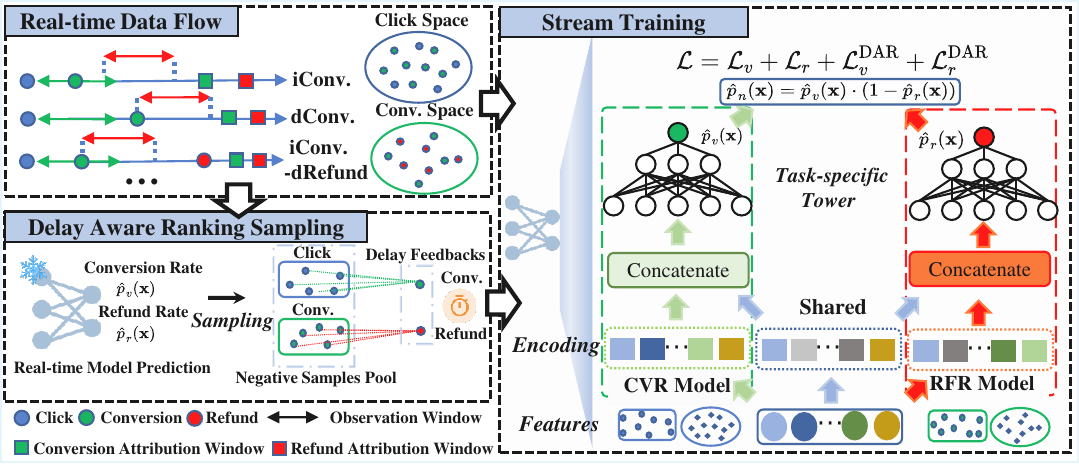}
\caption{Framework of \OursMethod. \textit{i} denotes an immediate response while
\textit{d} denotes a delayed response. }
\label{fig:framework}
\end{figure*}

\subsubsection{The Significance of Online Continuous Training.}\label{sec:analyze_stream}

To analyze intra-day patterns of NetCVR targets, we randomly sampled three days' data and computed hourly CVR and NetCVR. CVR was derived by aggregating clicks within each hour and assigning conversion labels based on user behavior over the following 3-day window (conversion attribute window). NetCVR was calculated similarly, with refund labels assigned based on user behavior over a 6-day window (conversion and refund attribute window).  

We can observe from Fig.~\ref{fig:hourly_cvr_netcvr_std} that both CVR and NetCVR display hour-to-hour fluctuations over time. Notably, both indicators experience a substantial decline during late-night periods (e.g., 00:00–04:00). 
Consequently, relying on offline methods that update models only once per day~\cite{zhao2023entire} is insufficient to capture such fine-grained temporal patterns, highlighting the necessity of employing online continuous learning algorithms that can adapt to dynamic user behavior and provide more accurate real-time responses.

\begin{table}[t]
\caption{Net conversion rate performance between the cascade modeling approach and direct modeling approach.}
\label{tab:cascade_structure-vs-single_structure}
\centering
\begin{tabular}{@{} l c c c c@{}}
\toprule
\textbf{Structure} & \textbf{AUC$\uparrow$} & \textbf{NLL$\downarrow$} & \textbf{PCOC} & \textbf{PRAUC$\uparrow$} \\
\midrule
Cascade & 0.7525 & 0.1803 & 1.0393 & 0.1585 \\
 Direct  & 0.7514 & 0.1814 & 1.2367 & 0.1567 \\
\bottomrule
\end{tabular}
\end{table}

\subsubsection{The Superiority of Cascade Structure}\label{sec:analyze_divide}
We first grouped items by CVR and analyzed their relationships with RFR and NetCVR. 
As shown in Fig.~\ref{fig:cvr_rfr_corr}, we can observe that:  
(1) Items with higher CVR tend to have higher NetCVR, suggesting a possible positive correlation and indicating \textbf{potential for joint modeling of CVR and NetCVR with shared parameters}. 
(2) Items with higher CVR generally show lower RFR. However, RFR increases in the top CVR bucket $(0.8, 1]$, which reveals that \textbf{conversion and refund are mostly related but retain some distinct characteristics}.

There are two approaches exist for online continuous NetCVR modeling: (1) direct modeling NetCVR, and (2) cascaded modeling CVR and RFR. 
Based on above insights, we can use a dual-tower structure to implement both methods. 
The direct method predicts CVR and NetCVR tasks simultaneously, while the cascaded method predicts CVR and RFR tasks together. 

To evaluate these approaches for online NetCVR prediction, we extended the simplest online continual learning method FNC~\cite{gu2021real, chen2022asymptotically} to NetCVR task as there is no prior continuous methods for NetCVR and FNC's straightforward debiasing strategy can minimize experimental interference. 
The detail of FNC is shown in Appendix\ref{apx:baselines}. 
In this adapted method, conversion events enter the model through the native data flow, and refund events are input when actual refunds occur. 
We compared AUC, NLL, PRAUC, and PCOC for both variants as proxy metrics to evaluate the performance of netCVR prediction. 

As shown in Tab.~\ref{tab:cascade_structure-vs-single_structure}, the cascaded approach attains higher AUC and PR-AUC, signaling superior model ranking accuracy. AUC evaluates the model's ability to rank true conversions above non-conversions and refunds. 
(1) Direct modeling NetCVR cannot distinguish between non-conversions and refunds, often overfitting conversion intent features, leading to biased predictions. Thus, its PCOC values exceeding 1. 
(2) In contrast, a cascaded approach decomposes NetCVR into sequential sub-tasks, i.e., conversion prediction followed by refund prediction, allowing the model to \textbf{capture distinct behavioral patterns at each stage and produce more accurate, interpretable estimates}.

\begin{figure}[t]
\centering
\includegraphics[width=1\columnwidth]{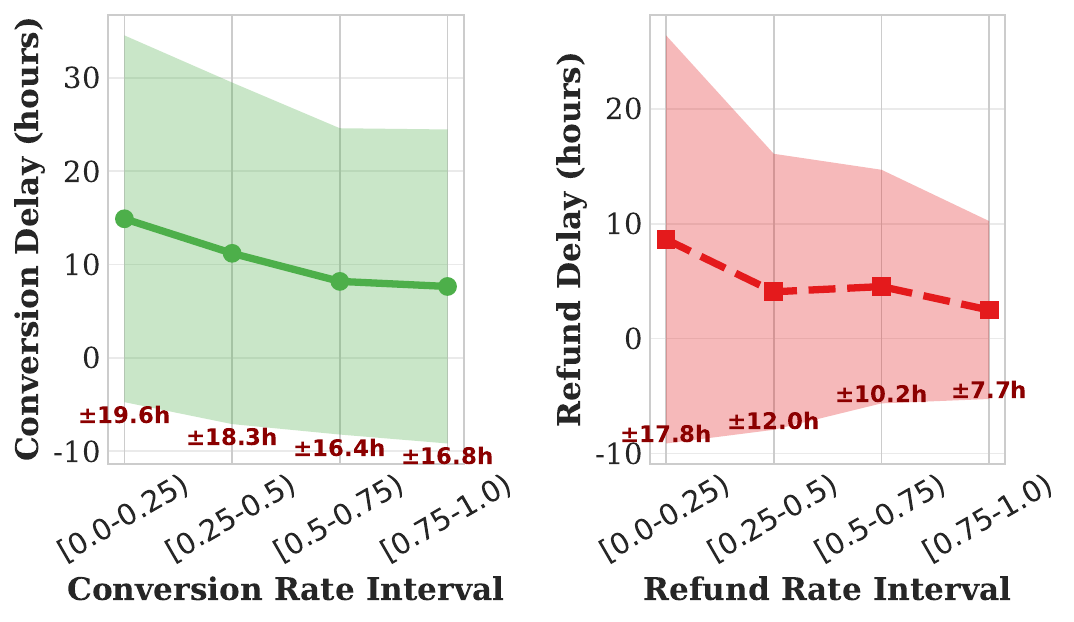} 
 %
\caption{Average conversion and refund delay across CVR and refund rate groups. 
}
\label{fig:delay_trend}
\end{figure}
\subsubsection{The Potential of Modeling Delay Time for NetCVR Prediction.} \label{sec:analyze_delay}
We grouped users by CVR and RFR and calculated the mean and variance of delay time for each group. 
We can observe from Fig.~\ref{fig:delay_trend} that: 
(1) High-CVR users have shorter and more consistent payment delays than low-CVR users, suggesting that those who convert quickly may have stronger purchase intent. 
(2) Compared to the low-RFR group, the high-RFR group has a shorter average delay and smaller variance, which may indicate immediate post-purchase dissatisfaction or that these users are subject to automated return policies.
\textbf{Delay is not just a technical challenge but a behavioral signal reflecting user confidence and satisfaction}. It should be considered in modeling cascaded delayed feedback.

    

\section{Methodology} \label{sec:Methodology}


Based on the gathered insights, we propose ne\acronym{T} conv\acronym{E}rsion ca\acronym{S}caded mode\acronym{L}ing and debi\acronym{A}sing (\OursMethod). 
As shown in Fig.~\ref{fig:framework}, \OursMethod consists of three components:
(1) Inspired by the insights that NetCVR exhibits clear temporal patterns (\S~\ref{sec:analyze_stream}), \OursMethod adopts an online continuous learning method with two-stage cascaded observation windows to dynamically track net conversion status (\S~\ref{sec:data_stream});
(2) Drawing on the insights that cascaded modeling of CVR and RFR for NetCVR is more effective, and that CVR and RFR are related but have distinct characteristics (\S~\ref{sec:analyze_divide}), \OursMethod employs a partially shared architecture to jointly model the CVR and RFR cascade ((\S~\ref{sec:Structure}). 
We apply a stage-wise debiasing strategy to keep accuracy (\S~\ref{sec:debiasing});
(3) Further guided by insight that delay time reflects user intent and satisfaction (\S~\ref{sec:analyze_delay}), \OursMethod adopts a \textbf{Delay-Aware Ranking loss} to adaptively adjust pairwise loss and emphasize reliable interactions (\S~\ref{sec:dabpr}).







\subsection{Data Stream} \label{sec:data_stream}

\begin{figure}[t]
\centering
\includegraphics[width=1\columnwidth]{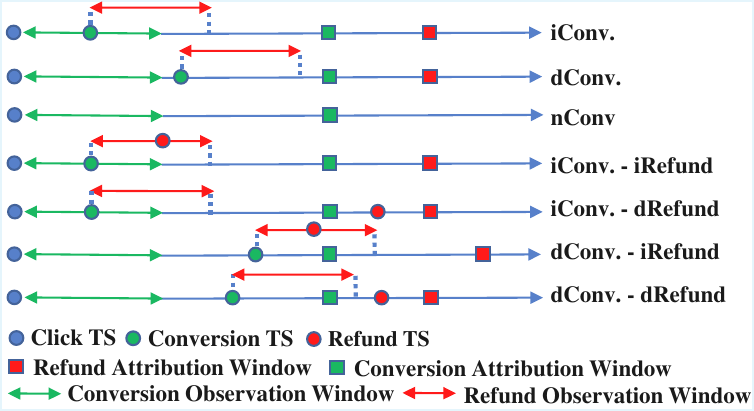}
\caption{The data flow of \OursMethod. 
\textit{i} denotes an immediate response while
\textit{d} denotes a delayed response.}
\label{fig:data_stream}
\end{figure}

In the complex landscape of NetCVR, traditional CVR prediction data streams fall short due to the multi-stage delayed feedback. 
To address this, we pioneer a novel stream data structure tailored for net conversion. 
As illustrated in Fig.~\ref{fig:data_stream}, \OursMethod establishes a structured pipeline with a dedicated \textbf{refund observation window} triggered post-conversion. 
This secondary window, analogous to the post-click \textit{conversion observation window}, enables early refund signal detection, ensuring accurate and timely NetCVR estimation.

\OursMethod further extends to classifying user interactions into seven trajectory types based on timing patterns. These types can be grouped into two overarching categories:
(1) \textbf{Conversion with No Refund}, including 
immediate conversion (\textbf{iConv.}), 
delayed conversion (\textbf{dConv.}), 
and no conversion (\textbf{nConv.}). 
(2) \textbf{Conversion with Refund}, including 
immediate conversion with immediate refund (\textbf{iConv.-iRefund}) and with delayed refund (\textbf{iConv.-dRefund}); 
delayed conversion with immediate refund (\textbf{dConv.-iRefund}) and with delayed refund (\textbf{dConv.-dRefund}). 

This refined classification allows the model to effectively differentiate between transient and sustained conversions, which is pivotal for enhancing the accuracy of net conversion estimation and supporting robust debiasing and dynamic label modeling in cascaded delays.

\subsection{CVR-RFR Cascaded Structure} \label{sec:Structure}

According to \S~\ref{sec:analyze_divide}, direct NetCVR modeling overlooks the dynamic process of $\text{Conversion} \to \text{Refund}$, leading to poor real-world calibration.
Thus, \textit{NetCVR should be modeled via a cascade framework that integrates CVR and RFR}.
Given that CVR and RFR are related but have distinct characteristics (\S~\ref{sec:analyze_divide}), the model should capture both their shared and unique features.

Inspired by Progressive Layered Extraction~\cite{2020Progressive}, we use a shared-private architecture to jointly learn CVR and RFR while reducing task interference. 
\OursMethod includes two task-private embedding layers $\embeddings_{\conversion}, \embeddings_{\refund}$ and one task-shared encoding module $\encodings_{\share}$. 
The private embedding layers model conversion and refund differences while the shared module transfer common knowledge, capturing the relation between the two. 
After encoding, \OursMethod combines shared and private information and feeds it into task-specific heads to predict 
$\hat{\pro}_{\conversion}(\feature)$ and $\hat{\pro}_{\refund}(\feature)$.


\subsection{Debiasing Strategies}
\label{sec:debiasing}


The cascaded delayed nature of NetCVR introduces system bias into the observed data, affecting accuracy.
To achieve an unbiased NetCVR estimate, we uses stage-wise importance weighting to debias CVR and RFR: first addressing click-to-conversion delay, then conversion-to-refund delay.

Both the CVR and RFR towers are challenged by delayed labels, necessitating a debiasing approach for accurate estimation. 
In both towers, a sample is positive only if the event occurs within the observation window; otherwise, it is initially negative. 
Delayed feedback creates censored positives, which are corrected via a recapture stream upon late observation. 
However, this delayed correction introduces cascaded delay bias, causing prolonged label uncertainty and temporally misaligned gradients.


To address censored delayed labels in both towers, we employ importance weighting~\cite{yang2021capturing} to reweight positive and negative samples. Detailed derivations are provided in the Appendix~\ref{apx:Debias_ESDFM}.
The importance weights for the CVR tower are defined as:
\begin{small}
    \begin{align}
    \weight_\conversion^{+}(\feature) &= 1 + \pro_{\conversion}(\feature) \cdot \pro(\delay_\conversion > \window_\conversion^{\text{obs}} \mid \conLabel=1, \feature), \\
    \weight_\conversion^{-}(\feature) &= \frac{(1 - \pro_{\conversion}(\feature)) \cdot \left(1 + \pro_{\conversion}(\feature) \cdot \pro(\delay_\conversion > \window_\conversion^{\text{obs}} \mid \conLabel=1, \feature)\right)}{1 - \pro_{\conversion}(\feature) + \pro_{\conversion}(\feature) \cdot \pro(\delay_\conversion > \window_\conversion^{\text{obs}} \mid \conLabel=1, \feature)}, 
\end{align}
\end{small}
where $\weight^{+}$ and $\weight^{-}$ represents the weights for positive and negative samples, 
$\pro_\conversion$ is the conversion probability estimated by the online streaming model.
$\pro(\delay_\conversion > \window_\conversion^{\text{obs}} \mid \conLabel=1, \feature)$ is the probability (estimated by a pretrained delay distribution model trained on historical logs follows prior approaches~\cite{yang2021capturing}) that a true conversion is delayed beyond the observation window $\window_\conversion^{\text{obs}}$ and thus mislabeled. 
$\delay_\conversion = \TIME_\conversion - \TIME_\click$ represents the conversion delay time, 
$\window_\conversion^{\text{obs}}$ represents the conversion observation window, 
$\TIME_\conversion, \TIME_\click, \TIME_\observe$ denotes conversion time, click time and observation time, respectively. 
$\posSet_\conversion$ and $\negSet_\conversion$ denote the sets of real-time observed positive and negative conversion samples respectively, which can be estimated via a survival model or empirical statistics. 
For the RFR tower, analogous importance weights are applied:

\begin{small}
    \begin{align} 
    \weight_\refund^+(\feature) &= 1 + \pro_{\refund}(\feature) \cdot \pro(\delay_{\refund} > \window_\refund^{obs} \mid \refLabel=1, y=1, \feature), \\
    \weight_\refund^-(\feature) &= \frac{(1 - \pro_{\refund}(\feature)) \cdot \left(1 + \pro_{\refund}(\feature) \cdot \Pr(\delay_{\refund} > \window_\refund^{obs} \mid \refLabel=1, y=1, \feature)\right)}{1 - \pro_{\refund}(\feature) + \pro_{\refund}(\feature)\cdot \Pr(\delay_{\refund} > \window_\refund^{obs} \mid \refLabel=1, y=1, \feature)},
\end{align}
\end{small}
where $\pro_{\refund}$ denotes the refund probability estimated by the online streaming model.
$\pro(\delay_{\refund} > \window_\refund^{obs} \mid \refLabel=1, y=1, \feature)$ represents the probability that a true refund ($\refLabel=1$) is delayed beyond the observation window ($\delay_{\refund} > \window_\refund^{obs}$). It is also estimated by a delay distribution model trained on historical logs.
$\delay_\refund = \TIME_\refund - \TIME_\conversion$ represents the refund delay time, 
$\window_\refund^{obs}$ represents the refund observation window, 
$\TIME_\refund$ denotes refund time. 
$\posSet_\refund$ and $\negSet_r$ denote the observed positive and negative refund samples among converted instances. 

The debiased loss functions for both towers are formulated as follows. For the CVR tower: 
\begin{small}
\begin{equation}
    \loss_{\conversion} = - \sum_{(\feature, \conLabel) \in \posSet_\conversion \cup \negSet_\conversion} 
    \big[
     y \cdot \weight_\conversion^+(\feature) \cdot \log \hat{\pro}_\conversion(\feature)
    + (1 - y) \cdot \weight_\conversion^-(\feature) \cdot \log\big(1 - \hat{\pro}_\conversion(\feature) \big)
    \big],
    \end{equation}
\end{small}
For the RFR tower, 
\begin{small}
\begin{equation}
    \loss_\refund = -\sum_{(\feature, \refLabel) \in \posSet_\refund \cup \negSet_\refund} \big[
     \refLabel \cdot \weight_\refund^+(\feature) \cdot \log \hat{\pro}_\refund(\feature) 
    + (1 - \refLabel) \cdot \weight_\refund^-(\feature) \cdot \log \big( 1 - \hat{\pro}_\refund(\feature) \big)
    \big],
\end{equation}
\end{small}

This two-stage debiasing framework ensures asymptotically unbiased NetCVR estimation by enabling $\hat{\pro}_\conversion(\feature)$ and $\hat{\pro}_\refund(\feature)$ to approach their respective ground-truth expectations $\hat{\pro}_{\net}(\feature) = \hat{\pro}_{\conversion}(\feature) \cdot (1 - \hat{\pro}_\refund(\feature))$. 
By stage-based bias correction, \OursMethod effectively manages the dual-delay structure in cascaded feedback, enhancing the applicability of single-stage debiasing techniques to the intricate NetCVR setting.

\subsection{Delay-Aware Ranking Loss}
\label{sec:dabpr}
Online CVR and NetCVR models are typically trained by point-wise loss. 
However, delayed feedback, especially the multi-stage cascaded delays in NetCVR prediction, can cause label uncertainty and potential label flipping. 
Point-wise loss struggles to capture this uncertainty, limiting its effectiveness in real-world scenarios.

To address this, we introduce pair-wise loss, which emphasizes ranking positive samples higher than negative ones. However, the inconsistent quality of samples can undermine its effectiveness. 
Fortunately, our analysis in Section~\ref{sec:analyze_delay} reveals a relationship between delay time and conversion/refund behavior, which provides a solution.  
By leveraging delay time as a signal, we can increase the weight of high-quality positive samples and select more reliable negative samples.

\subsubsection{Delay-Based Positive Sample Weighting} 
We employ a reweighting approach to differentially treat positive samples. 
For a batch of observed positive conversion samples $\posSet_\conversion$, let $\mathbf{\delay_\conversion} = \{\delay_\conversion^{(i)}\}_{i \in \posSet_\conversion}$ denote their corresponding conversion delays, where $\delay_\conversion^{(i)} = \TIME_\conversion^{(i)} - \TIME_\click^{(i)}$ is the delay time of each sample, $\TIME_\conversion, \TIME_\click$ are the conversion time and click time. 
Since \textit{high-CVR items convert faster with smaller delay variance} (\S~\ref{sec:analyze_delay}), we assign each positive sample a weight based on its delay:
\begin{small}
    \begin{equation}
    \weight_i = \weight(\delay_\conversion^{(i)}) = \weight_{\min} + \alpha \cdot \sigma \big( ({m - \delay_\conversion^{(i)}})/{s} \big),
\end{equation}
\end{small}
where $m = \mathrm{median}(\mathbf{h}_\conversion)$ is the median conversion latency in the batch, 
$s = \max(\mathrm{std}(\mathbf{h}_\conversion), 1)$ measures delay dispersion. We clamp $s$ to prevent overconfidence due to small variance.
$\sigma(\cdot)$ is the sigmoid function and $\weight_{\min} > 0$ to ensures slow converters still contribute learning signals.
$\alpha$ is a hyper-parameter. 
This design allocates greater weights to fast conversions samples. 

\subsubsection{Uncertainty-Based Negative Sample Sampling} 
We replace the random negative sampling process in ranking tasks. Here, we assume that the lower the predicted CVR, the higher the likelihood of a sample being a true negative. 
we define the sampling probability and normalize it as follows: 
\begin{small}
\begin{equation}
    \pi_j = \frac{\big( 1 - \hat{\pro}_\conversion(\feature_j) \big) / \tau}{\sum_{k \in \negSet_\conversion} [\big(1 - \hat{\pro}_\conversion(\feature_k)\big) / \tau]}, \quad j \in \negSet_\conversion,
\end{equation}
\end{small}
where $\tau > 0$ is a temperature hyperparameter that controls the softness of the sampling process. 
$\hat{\pro}_\conversion$ is the conversion probability, $\pi$ is the sample probability. 
We then independently sample $K$ negative items for each positive item $i \in \posSet_\conversion$ according to the distribution $\pi$, forming $K$ pairs $\{ (i, j_1), \dots, (i, j_K) \}$.

\subsubsection{Delay-aware Ranking Loss}


The loss for the conversion stage is calculated as:
\begin{small}
    \begin{equation}
        \loss^{\text{DAR}}_\conversion = \sum_{i \in \posSet_\conversion} \weight_i \cdot \left( -\frac{1}{K} \sum_{j \in \negSet_{\conversion}, k=1}^{K} \log \sigma(o^{\text{CVR}}_{i} - o^{\text{CVR}}_{j_k}) \right),
    \end{equation}
\end{small}
where $\weight_i$ is the weight for positive sample $i$, $o^{\text{CVR}}_i = f_\theta^{\text{CVR}}(\feature_i)$ denote the output for positive sample $i$, and $\{o_{j}\}_{k=1}^K$ denote the logit outputs of its sampled negatives. 
This loss function combines delay-based positive weighting and uncertainty-aware negative sampling, enabling more accurate and stable preference learning under conversion delay.
Similarly, we apply the same approach to refunds,  
\begin{small}
    \begin{equation}
        \loss^{\text{DAR}}_\refund = \sum_{i \in \posSet_\refund} \weight_i \cdot \left( -\frac{1}{K} \sum_{j \in \negSet_{\refund}, k=1}^{K} \log \sigma(o_i^{\text{RFR}} - o_{j_k}^{\text{RFR}}) \right),
    \end{equation}
\end{small}
where key conversion-related variables are substituted with their refund counterparts, $\posSet_\conversion, \negSet_\refund$ are the positive and negative refund sample set, 
$o^{\text{RFR}}_i = f_\theta^{\text{RFR}}(\feature_i)$ denote the output for sample $i$.
This highlights the symmetric and versatile nature of Delay-aware Ranking Loss in handling both stages of cascaded delayed feedback.

The final loss consists of the debiased CVR loss, debiased RFR loss, delay-aware CVR ranking Loss and delay-aware RFR ranking Loss, 
\begin{small}
\begin{equation}
\label{equ:final_loss}
\loss = \loss_{\conversion}  + \loss_{\refund} +  \loss^{\text{DAR}}_\conversion + \loss^{\text{DAR}}_\refund,
\end{equation}
\end{small}
\begin{table*}[t]
\centering
\caption{Performance Comparison on CVR and NetCVR prediction. 
Oracle is the upper bound on the best achievable performance as there are no delayed Conversions. 
"\textbf{Shared}", "\textbf{Separate}", and "\textbf{Hybrid}" are three variants of \OursMethod. 
\textbf{Bold} indicates the best results and the oracle results for reference are \textcolor{gray}{grayed out}.
}

\label{tab:main_exp_results}
\resizebox{0.96\textwidth}{!}{%
\begin{tabular}{@{} l c c c c c @{}}  
\toprule
\textbf{Model} 
& \makecell{\textbf{AUC}$\uparrow$ \\ \textbf{(CVR / NetCVR)}} 
& \makecell{\textbf{NLL}$\downarrow$ \\ \textbf{(CVR / NetCVR)}} 
& \makecell{\textbf{PRAUC}$\uparrow$ \\ \textbf{(CVR / NetCVR)}} 
& \makecell{\textbf{RI-AUC}$\uparrow$ \\ \textbf{(CVR / NetCVR)}} 
& \makecell{\textbf{RI-PRAUC}$\uparrow$ \\ \textbf{(CVR / NetCVR)}} \\
\midrule

\textbf{Pre-trained}
    & 0.7223 / 0.7353 & 0.2252 / 0.1810 & 0.1768 / 0.1521 & $\sim$ & $\sim$ \\

\textcolor{gray}{\textbf{Oracle}}
   & \textcolor{gray}{0.7556 / 0.7643} & \textcolor{gray}{0.2155 / 0.1750} & \textcolor{gray}{0.2007 / 0.1722} & \textcolor{gray}{$\sim$} & \textcolor{gray}{$\sim$} \\

\textbf{BDL} 
    & 0.7381 / 0.7476 & 0.2237 / 0.1805 & 0.1832 / 0.1574 & 47.45\% / 42.41\% & 26.78\% / 26.37\% \\

\textbf{FNC}
    & 0.7437 / 0.7525 & 0.2225 / 0.1801 & 0.1852 / 0.1585 & 64.26\% / 59.31\% & 35.15\% / 31.84\% \\

\textbf{FNW (RecSys'19)}
    & 0.7443 / 0.7539 & 0.2220 / 0.1800 & 0.1887 / 0.1619 & 66.07\% / 64.14\% & 49.79\% / 48.76\% \\

\textbf{DEFER (KDD'21)}
    & 0.7450 / 0.7552 & 0.2190 / 0.1777 & 0.1920 / 0.1649 & 68.17\% / 68.62\% & 63.60\% / 63.68\% \\

\textbf{DEFUSE (WWW'22)}
    & 0.7459 / 0.7550 & 0.2347 / 0.1971 & 0.1925 / 0.1645 & 70.87\% / 67.93\% & 65.69\% / 61.69\% \\

\textbf{ES-DFM (AAAI'21)}
    & 0.7456 / 0.7546 & 0.2190 / 0.1783 & 0.1924 / 0.1644 & 69.97\% / 66.55\% & 65.27\% / 61.19\% \\

\textbf{DSFN (SIGIR'23)}
    & 0.7437 / 0.7527 & 0.2199 / 0.1793 & 0.1905 / 0.1617 & 64.26\% / 60.00\% & 57.32\% / 47.76\% \\

\textbf{DDFM (CIKM'23)}
    & 0.7446 / 0.7541 & 0.2199 / 0.1793 & 0.1918 / 0.1640 & 66.97\% / 64.83\% & 62.76\% / 59.20\% \\

\textbf{MISS (AAAI'24)}
    & 0.7441 / 0.7527 & 0.2199 / 0.1783 & 0.1902 / 0.1625 & 65.47\% / 60.00\% & 56.07\% / 51.74\% \\
\midrule

\textbf{\OursMethod (Separate)}  
    & 0.7456 / 0.7555 & 0.2190 / 0.1776 & 0.1924 / 0.1651 & 69.97\% / 69.66\% & 65.27\% / 64.68\% \\

\textbf{\OursMethod (Shared)}  
   & 0.7458 / 0.7567 & 0.2189 / 0.1772 & 0.1926 / 0.1667 & 70.57\% / 73.79\% & 66.11\% / 72.64\% \\

\textbf{\OursMethod (Hybrid)}  
& 0.7469 / 0.7579 & 0.2188 / 0.1771 & 0.1933 / 0.1676 & 73.87\% / 77.93\% & 69.04\% / 77.11\% \\

\textbf{\OursMethod} 
    & \textbf{0.7479 / 0.7588} & \textbf{0.2182 / 0.1767} & \textbf{0.1939 / 0.1679} & \textbf{76.88\% / 81.03\%} & \textbf{71.55\% / 78.61\%} \\
\bottomrule
\end{tabular}%
}
\end{table*}

\section{Experiments}
\label{sec:exp}

In this section, we study the following research questions:
\textbf{RQ1:} Can \OursMethod improve the performance of net conversion rate prediction? (\S~\ref{sec:main_analysis})
\textbf{RQ2:} How does each component of \OursMethod affect the performance? 
Specifically, we perform ablation study on "Cascaded structure" (\S~\ref{sec:ablation-structure}), 
"Debiasing strategies" (\S~\ref{sec:ablation-debias}), 
and "Delay-aware ranking loss" (\S~\ref{sec:ablation-loss}). 
\textbf{RQ3:} How does observation window settings affect the search performance? (\S~\ref{sec:ablation_window})

\subsection{Experimental Setting}
\subsubsection{Data Settings \& Sample Delivery}
To simulate streaming training, we follow the online-learning paradigm~\cite{gu2021real,chen2022asymptotically}, split the data into pre-train and streaming chunks, and design a four-window mechanism together with a corresponding sample-delivery rule. Further details are given in Appendix~\ref{apx:data-settings} and Appendix~\ref{apx:sample-delivery}.

\subsubsection{Evaluation Metrics}
\label{sec:metrics}
We use AUC, PRAUC, NLL, RI-AUC, and RI-PRAUC to evaluate  NetCVR prediction performance~\cite{gu2021real, chen2022asymptotically, liu2024online}.
The detailed metrics are shown in Appendix~\ref{apx:metrics}.

\subsubsection{Competitors}
\label{sec:baselines}
We compare against three reference variants: 
\textbf{Pre-trained}, a static model without online updates; 
\textbf{Oracle}, a privileged model trained with complete feedback after full observation; and 
\textbf{BDL}, an offline model updated daily using complete labels after the attribution window. 
In addition, we benchmark \OursMethod against eight strong delayed-feedback baselines, including label-correction methods (FNC, FNW, ES-DFM, DEFER, DEFUSE) and hybrid multi-window approaches (DDFM, DSFN, MISS). 
Detailed descriptions are provided in Appendix~\ref{apx:baselines}.

\subsection{Performance Comparison} \label{sec:main_analysis}
We evaluate all methods on the \OursDataset dataset.  
We have following observations from Tab.~\ref{tab:main_exp_results}: 
(1) \OursMethod consistently enhances both CVR and NetCVR tasks. 
It outperforms other methods across all metrics, showing its effectiveness in tackling delayed feedback and better capturing user preferences. 
The improvements of RI-AUC in the NetCVR task (12.41\%) are more pronounced than those in the CVR task (6.01\%), suggesting that it is \textbf{more effective in addressing the multi-stage cascaded delayed feedback}. 
(2) \textbf{Continuous training contributes to the performance lift}. All online models perform better than BDL models, which validates the need for online continuous learning (\S~\ref{sec:analyze_stream}) and demonstrates the advantage of online continuous learning models in promptly updating user preferences for better recommendations.
(3) Adding Delay-aware Ranking Loss proves effective. Compared with \OursMethod (Hybird), \OursMethod improves RI-AUC by 3.01\% in CVR and 3.1\% in NetCVR. This shows that \textbf{using delay time helps online models better identify delayed labels}. 


\subsection{Ablation Study}

\subsubsection{Impact of Cascaded Structure} \label{sec:ablation-structure}
To explore the impact of cascaded structure, we have three variants:
(1) Fully shared model (\textbf{Shared}): CVR and RFR towers share all lower-layer parameters.
(2) Fully separate model (\textbf{Separate}): CVR and RFR towers have separate lower-layer parameters.
(3) Hybrid shared-independent model (Our method, \textbf{Hybrid}): CVR and RFR towers combine shared and task-specific lower-layer parameters.
We remove the delay-aware ranking loss to eliminate potential interference.

From Tab.~\ref{tab:main_exp_results} we can observe that: 
(1) \textbf{Hybrid} achieves the best results. By separating shared and unique parameters, \textbf{Hybrid} effectively balances knowledge transfer and task specificity. 
This validates the findings in \S~\ref{sec:analyze_divide}, which highlight that CVR and RFR share relevant information yet retain unique content, making this approach crucial for accurate NetCVR estimation.
(2) \textbf{Shared} performs better than \textbf{Separate}. Given the limited availability of refund data, parameter sharing enhances the performance of the RFR model, surpassing the results of completely independent modeling.

\subsubsection{Impact of Refund De-biasing}\label{sec:ablation-debias}
To explore the impact of refund debiasing, we conduct ablation studies by removing the Refund Observation Window (ROW) and debiasing strategy (DS) under \textbf{Separate}. Note that DS can only be used with ROW enabled. 

We can observe from Tab.~\ref{tab:ablation_refund_db} that: 
(1) Adding ROW alone enhances AUC and reduces NLL, demonstrating improved ranking accuracy and lower uncertainty. This indicates that \textbf{the observation window effectively filters early refunds}, thereby reducing label noise.
(2) Combining both ROW and DS achieves optimal performance. This suggests that while early filtering with ROW is beneficial, \textbf{an explicit debiasing strategy is indispensable} for precise NetCVR estimation.

\begin{table}[t]
\centering
\caption{
Ablation study on the Refund Observation Window (ROW) and Refund Debiasing Strategy (DS), built upon the \textbf{Separate} architecture in \S~\ref{sec:ablation-structure}. 
}
\label{tab:ablation_refund_db}
\begin{tabular}{c c c c c}
\toprule
\textbf{ROW} & \textbf{DS} & \textbf{AUC$\uparrow$} & \textbf{NLL$\downarrow$} & \textbf{PRAUC$\uparrow$} \\
\midrule
     $\times$  & $\times$  & 0.7546 & 0.1783 & 0.1644 \\
     $\checkmark$ & $\times$  & 0.7553 & 0.1779 & 0.1644 \\
     $\checkmark$ & $\checkmark$ & 0.7555 & 0.1776 & 0.1651 \\
\bottomrule
\end{tabular}
\end{table}



\begin{table}[t]
\centering
\caption{Ablation study on Delay-Aware Ranking Loss (DAR), built upon the \textbf{Hybrid} architecture in \S~\ref{sec:ablation-structure}. NR represents no ranking loss. PW, RN, LN, and HN denote positive weighting, random, low-uncertainty, and high-uncertainty negative sampling, respectively.}  
\label{tab:ablation_dabpr_hsp}
\resizebox{\columnwidth}{!}{%
\begin{tabular}{@{}l c c c c c@{}}
\toprule
\textbf{Variation} 
& \textbf{RI-AUC$\uparrow$} 
& \textbf{NLL$\downarrow$} 
& \textbf{RI-PRAUC$\uparrow$} \\
& \textbf{(CVR / NetCVR)} 
& \textbf{(CVR / NetCVR)} 
& \textbf{(CVR / NetCVR)} \\
\midrule

\textbf{NR}            & 73.87\% / 77.93\% & 0.2188 / 0.1771 & 69.04\% / 77.11\% \\
\textbf{RN}    & 75.68\% / 79.77\% & 0.2186 / 0.1767 & 69.60\% / 77.28\% \\
\textbf{HN}     & 75.08\% / 76.44\% & 0.2226 / 0.1801 & 69.60\% / 76.29\% \\
\textbf{LN}        & 76.08\% / 80.23\% & 0.2182 / 0.1767 & 70.85\% / 78.23\% \\
\textbf{DAR (PW-LN)} & \textbf{76.78\% / 80.92\%} & \textbf{0.2182 / 0.1767} & \textbf{71.41\% / 78.77\%} \\
\bottomrule
\end{tabular}%
}
\end{table}

\begin{sloppypar}
\subsubsection{Impact of Delay-aware Ranking Loss}
\label{sec:ablation-loss}
To explore the effectiveness of delay-aware ranking loss (\textbf{DAR}), we assess its key components—delay-aware positive weighting (PW) and low-uncertainty-driven negative sampling (LN) within the \textbf{Hybrid} in \S~\ref{sec:ablation-structure}. 
The variants include: \textbf{NR} (no ranking loss), \textbf{RN} (random negatives with ranking loss), \textbf{HN} (high-uncertainty negative sampling), \textbf{LN} (low-uncertainty negative sampling), and \textbf{DAR (PW-LN)} (our method with positive weighting and low-uncertainty negatives).
\end{sloppypar}

From Tab.~\ref{tab:ablation_dabpr_hsp}, we observe that: 
(1) \textbf{DAR} performs best, as it reduces potential noise in both positive and negative samples. 
(2) Randomly negatives (\textbf{RN}) outperforms high-uncertainty negatives (\textbf{HN}) but underperforms low-uncertainty negatives (\textbf{LN}), underscoring the importance of uncertainty-guided negative sampling.
(3) \textbf{PW-LN} is better than \textbf{LN}, demonstrating the effectiveness of positive sample weighting and the value of focusing on small delay times to capture high-confidence user intent (\S~\ref{sec:analyze_delay}).
(4) \textbf{NR} performs the worst, indicating that pair-wise ranking loss can mitigate label uncertainty caused by delayed feedback (\S~\ref{sec:dabpr}).

\begin{figure}[t]
\centering
\includegraphics[width=1\columnwidth]{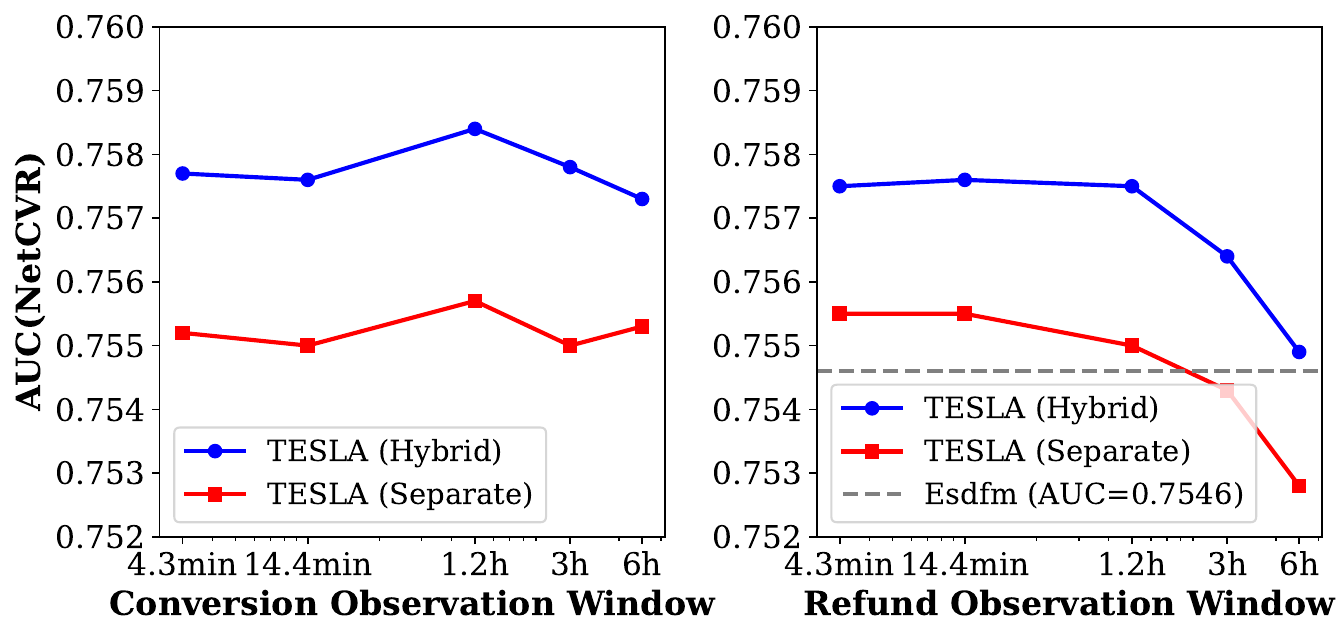}
\caption{Ablation study on refund and conversion observation window. 
}
\label{fig:window_ablation}
\end{figure}
\begin{figure}[t]
\centering
\includegraphics[width=1\columnwidth]{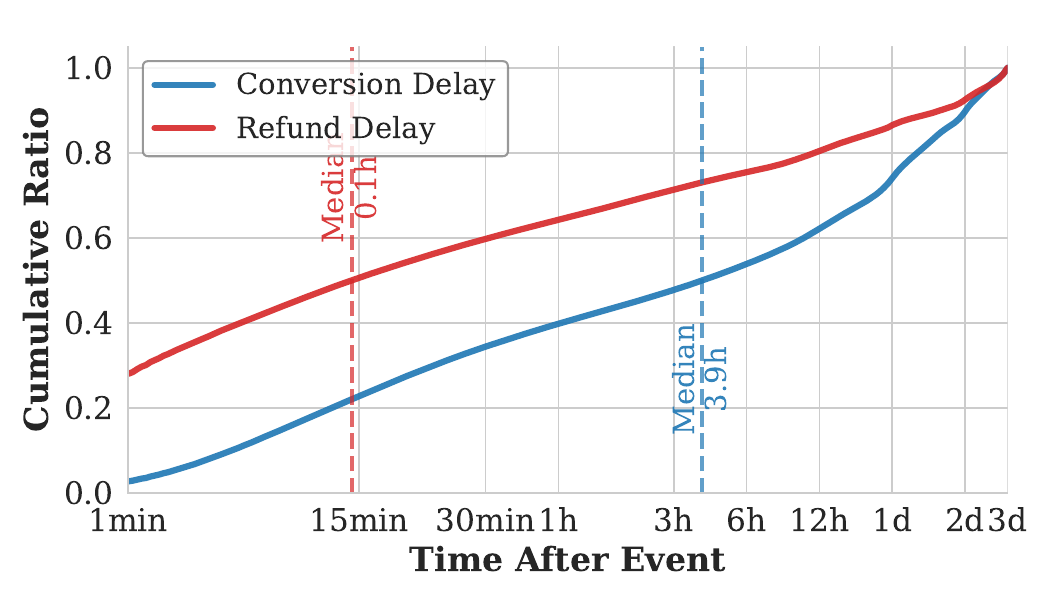}
\caption{Cumulative distribution of conversion and refund delays.} \label{fig:conversin_refund_delay_cdf}
\end{figure}

\subsection{Impact of Observation Windows}
\label{sec:ablation_window}

To optimize NetCVR performance through selecting and tuning the conversion and refund observation windows, we conduct parameter experiments on $\window^{obs}_\conversion$ and $W^{obs}_\refund$ over $\{0.003, 0.01, 0.05, 0.125, 0.25\}$ days, modifying one window at a time while fixing the other at 0.01 days. AUC (NetCVR) serves as the primary evaluation metric.

From Fig.~\ref{fig:window_ablation}, we can observe that: 
(1) Conversion observation window $\window^{obs}_\conversion$ are achieved at 0.05 days, with performance declining beyond this point. This highlights the need for a balance between sufficient time to gather stable positive signals and maintaining freshness, reflecting the slower and more diverse conversion dynamics as shown in Fig.~\ref{fig:conversin_refund_delay_cdf}.
(2) In contrast, shorter refund observation window $W^{obs}_r$ are preferable. 
AUC remains stable when $\window^{obs}_\refund \leq 0.01$ days and declines with longer durations. 
When the window is too long, the results may even be worse than those without an refund observation window.
This aligns with the refund delay distribution in Fig.~\ref{fig:conversin_refund_delay_cdf}, where most refunds occur quickly. 
\textbf{Shorter window enhances label purity by excluding delayed cases, improving training reliability.}

\section{Conclusion}
In this paper, We study multi-stage cascaded delayed feedback in NetCVR prediction and introduce \OursDataset, \textbf{the first large-scale open dataset for continuous NetCVR modeling} from Taobao. 
Analysis reveals key temporal patterns and the benefit of cascaded modeling. 
Based on these insights, we propose \OursMethod, \textbf{the first online continuous learning method addressing cascaded delayed feedback} with a CVR-RFR structure, stage-aware debiasing, and delay-aware ranking loss. 
Experiments show its superior performance over existing methods. 
We hope that our exploration will be helpful for future research in this area.

\section*{Acknowledgments}
This work was supported by the Natural Science Foundation of China (No.62432011,62372390), Science Fund for Distinguished Young Scholars of Fujian Province (No.2025J010001), and Alibaba Group through Alibaba Innovative Research Program. 
Chen Lin is the corresponding author. This work was completed while the first author was an intern at Alibaba.

\clearpage
\balance
\bibliographystyle{ACM-Reference-Format}
\bibliography{references}

\appendix
\section{Key Notations}\label{apx:notation}

The key notations used throughout this paper are summarized in Table~\ref{tab:notation}.

\begin{table}[htbp]
\centering
\caption{Notation and explanations}
\label{tab:notation}
\begin{tabularx}{0.45\textwidth}{@{} >{\bfseries}l X @{}}
\toprule
Notation & Explanation \\ \midrule
$\mathbf{x},y,z$ & Input features, conversion labels, refund labels \\
$t_c,t_v,t_r,t_o$ & click time, conversion time, refund time, observation time \\
$e_v,e_r$ & $e_v = t_o - t_c$ :interval between click and observation time; $e_r = t_o - t_v$ :interval between conversion and observation time \\
$h_v,h_r$ & conversion delay : $h_v = t_v - t_c$; refund delay : $h_r = t_r - t_v$ \\
$\window_\conversion^{obs},\window_\refund^{obs}$ & conversion/refund observation window\\
$\window_\conversion^{attr},\window_\refund^{attr}$ & conversion/refund Attribution window\\
$p_v(x),q_v(\mathbf{x})$ & ground-truth of conversion : $p_v(\mathbf{x}) = p(y=1|\mathbf{x})$, observerd distribution of conversion : $q_v(\mathbf{x}) = q(y=1|\mathbf{x})$\\
$p_r(\mathbf{x}),q_r(\mathbf{x})$ & ground-truth of refund : $p_r(\mathbf{x}) = p(z=1|y=1,\mathbf{x})$, observerd distribution of refund : $q_r(\mathbf{x}) = q(z=1|y=1,\mathbf{x})$\\
$p_n(\mathbf{x}),q_n(\mathbf{x})$ & ground-truth of net conversion : $p_n(\mathbf{x}) = p_v(x) \cdot (1 - p_r(\mathbf{x}))$, observerd distribution of net conversion : $q_n(\mathbf{x}) = q_c(\mathbf{x}) \cdot (1 - q_r(\mathbf{x})$\\

$\mathcal{P}_v,\mathcal{P}_r$ & real-time positive conversion/refund samples set from streaming data \\
$\mathcal{N}_v,\mathcal{N}_r$ & real-time negative conversion/refund samples set from streaming data \\
$\hat{p}_v(\mathbf{x}), \hat{p}_r(\mathbf{x}),\hat{p}_n(\mathbf{x})$ & output logits of the CVR and RFR models, i.e., $\hat{p}_v(\mathbf{x}) = f_{\theta}^{\text{CVR}}(\mathbf{x})$, $\hat{p}_r(x) = f_{\theta}^{\text{RFR}}(\mathbf{x}))$ , $\hat{p}_n(x) = \hat{p}_v(\mathbf{x}) \cdot (1 - \hat{p}_r(\mathbf{x}))$\\

\bottomrule
\end{tabularx}
\end{table}

\section{Dataset Construction}\label{apx:data_construction}
\OursDataset not only includes clicks and conversions but also incorporates refund information, along with timestamps for all interactions. 
These features allow for accurate modeling of NetCVR prediction with multi-stage cascaded delayed feedback and also provide benefits for CVR and RFR prediction tasks. 
The dataset covers sampled clicks in 25 days, and the max attribution window is 15 days. The statistics are summarized in Tab. ~\ref{tab:main_statistics}.

\subsection{Data Collection}
\OursDataset is collected from Taobao, one of the largest E-commerce sites worldwide, designed to support research on post-click value prediction under realistic online continuous learning conditions. 
To ensure behavioral diversity and representativeness, we sample from multiple advertising scenarios across different traffic and intent levels, covering both high-exposure placements and intent-rich contexts. Users are stratified by activity level and sampled to maintain balanced behavioral coverage over time, forming a stable, million-scale training set.
To reflect the characteristics of NetCVR prediction, we upsample active users exhibiting high refund rates. 
This dataset does not represent real-world business metrics or operational conditions.

\subsection{Feature Profile} 
Each ad click instance includes two types of signals:  
(1) \textit{Categorical features}, consisting of 22 fields in total: 8 user features (e.g., age groups and membership levels), 6 item features (e.g., category, brand), and 8 context features. All identifiers are irreversibly hashed to ensure privacy.  
(2) \textit{Temporal information}, including the click timestamp and the sequence of subsequent conversion and refund timestamps, which enable modeling of delayed feedback and dynamic user behavior.

\subsection{Data Compliance and Availability}
All personally identifiable and business-sensitive information, such as raw user, item, campaign, and placement IDs, is rigorously de-identified or removed. Only features essential for modeling are retained, and the dataset does not reflect real-time business operations. 
We will release \OursDataset as a public benchmark to facilitate research on netCVR prediction with privacy-preserving, real-world data.
\begin{table*}[th]
\centering
\caption{Statistics of the datasets. Counts are rounded in millions (M). }
\label{tab:main_statistics}
\begin{tabularx}{\textwidth}{@{}l *{10}{>{\centering\arraybackslash}X}@{}}
\toprule
\textbf{Dataset} & 
\textbf{\#Users} & 
\textbf{\#Items} & 
\textbf{\#Clicks} & 
\textbf{\#Conversions} & 
\textbf{CVR} & 
\textbf{\#Refunds} & 
\textbf{Duration} \\
\midrule
\OursDataset & 2.0M & 3.1M & 41.0M & 3.7M & 0.091 &  2.0M  & 25 days \\
\bottomrule
\end{tabularx}
\end{table*}

\section{Debiasing Strategy in ES-DFM}\label{apx:Debias_ESDFM}
In streaming conversion rate prediction, due to delayed feedback, the observed label distribution $ \obsPro(y|\feature) $ is biased. Positive conversions that occur after the observation time are mislabeled as negatives, leading to underestimation of the true conversion rate $ \realPro_\conversion(\feature) = \realPro(\conLabel=1|\feature) $. 

To achieve unbiased estimation, ES-DFM adopt importance sampling under the \emph{elapsed-time sampling} framework. The key idea is to model the elapsed time $ \interval_\conversion = t_o - t_v $ as a random variable with distribution $ \realPro(\interval_\conversion|\feature) $, and then correct for selection bias via reweighting.

Due to delayed feedback, only conversions with $ h_c \leq \interval_\conversion $ are observed. Hence, the observed distribution is:
\begin{align}
    \obsPro(\conLabel=1|\feature) &= \realPro(\conLabel=1|\feature) \cdot \realPro(\delay_\conversion \leq \interval_\conversion \mid \conLabel=1, \feature), \\
    \obsPro(\conLabel=0|\feature) &= \realPro(\conLabel=0|\feature) + \realPro(\conLabel=1|\feature) \cdot \realPro(\delay_\conversion > \interval_\conversion \mid \conLabel=1, \feature).
\end{align}

In ES-DFM, when a delayed conversion is later observed, It insert a positive duplicate into the training stream. This increases the total number of samples by $ \realPro(\conLabel=1|\feature) \cdot \realPro(\delay_\conversion > \interval_\conversion \mid \conLabel=1, \feature) $. Therefore, the observed distribution must be normalized by $ 1 + \realPro(\conLabel=1|\feature) \cdot \realPro(\delay_\conversion > \interval_\conversion \mid \conLabel=1, \feature) $, yielding:
\begin{align}
    \obsPro(\conLabel=1|\feature) &= \frac{\realPro(\conLabel=1|\feature)}{1 + \realPro(\conLabel=1|\feature) \cdot \realPro(\delay_\conversion > \interval_\conversion \mid \conLabel=1, \feature)}, \\
    \obsPro(\conLabel=0|\feature) &= \frac{\realPro(\conLabel=0|\feature) + \realPro(\conLabel=1|\feature) \cdot \realPro(\delay_\conversion > \interval_\conversion \mid \conLabel=1, \feature)}{1 + \realPro(\conLabel=1|\feature) \cdot \realPro(\delay_\conversion > \interval_\conversion \mid \conLabel=1, \feature)},
\end{align}

ES-DFM apply importance sampling to recover the expectation under the true distribution $ \realPro(y|\feature) $:
\begin{equation}
\begin{aligned}
    \mathcal{L}_{\text{ideal}} 
    &= \mathbb{E}_{(\feature,\conLabel) \sim \realPro(\feature,\conLabel)} \left[ \ell(\conLabel, \hat{\realPro}_\conversion(\feature)) \right], \\
    &= \mathbb{E}_{(\feature,\conLabel) \sim \obsPro(\feature,\conLabel)} \left[ \frac{\realPro(\conLabel|\feature)}{\obsPro(\conLabel|\feature)} \cdot \ell(\conLabel, \hat{\realPro}_\conversion(\feature)) \right],
\end{aligned}
\end{equation}

The importance weights are computed as follows:

\paragraph{For positive samples ($y = 1$):}
\begin{equation}
\begin{aligned}
    \weight_\conversion^+ 
    &= \frac{\realPro(\conLabel=1|\feature)}{\obsPro(\conLabel=1|\feature)}, \\
    &= \frac{\realPro(\conLabel=1|\feature) \cdot (1 + \realPro(\conLabel=1|\feature) \cdot \realPro(\delay_\conversion > \interval_\conversion \mid \conLabel=1, \feature))}{\realPro(\conLabel=1|\feature)}, \\
    &= 1 + \realPro(\conLabel=1|\feature) \cdot \realPro(\delay_\conversion > \interval_\conversion \mid \conLabel=1, \feature),
\end{aligned}
\end{equation}

\paragraph{For negative samples ($y = 0$):}
\begin{equation}
\begin{aligned}
    \weight_\conversion^- 
    &= \frac{\realPro(\conLabel=0|\feature)}{\obsPro(\conLabel=0|\feature)}, \\
    &= \frac{\realPro(\conLabel=0|\feature)}{\dfrac{\realPro(\conLabel=0|\feature) + \realPro(\conLabel=1|\feature) \cdot \realPro(\delay_\conversion > \interval_\conversion \mid \conLabel=1, \feature)}{1 + \realPro(\conLabel=1|\feature) \cdot \realPro(\delay_\conversion > \interval_\conversion \mid \conLabel=1, \feature)}}, \\
    &= \frac{\bigl(1 - \realPro(\conLabel=1|\feature)\bigr) \cdot \bigl(1 + \realPro(\conLabel=1|\feature) \cdot \realPro(\delay_\conversion > \interval_\conversion \mid \conLabel=1, \feature)\bigr)}{(1 - \realPro(\conLabel=1|\feature)) + \realPro(\conLabel=1|\feature) \cdot \realPro(\delay_\conversion > \interval_\conversion \mid \conLabel=1, \feature)},
\end{aligned}
\end{equation}

Using the binary cross-entropy loss and substituting the importance weights, we obtain the final loss:
\begin{small}
\begin{equation}
\mathcal{L}_{\text{ES-DFM}} = -\sum_{(\feature, \conLabel) \in \posSet_\conversion \cup \negSet_\conversion} \left[
\conLabel \cdot  \weight_\conversion^+ \cdot \log \hat{p}_v(\feature)
+ (1 - \conLabel) \cdot  \weight_\conversion^- \cdot \log(1 - \hat{p}_v(\feature))
\right],
\end{equation}
\end{small}

\section{Experiments}\label{apx:Exp}
\subsection{Data Settings}\label{apx:data-settings}
To simulate continuous training, we split our dataset into two parts.
The first part is for pre-training to provide a strong initial setup. 
We put a special gap between the two parts based on conversion and refund attribution windows ($\window^{attr}_\conversion + \window^{attr}_\refund$) to prevent label leakage. 
The second part is used to train and evaluate the models and is further divided into 0.01-day (15-minute) segments. 
Following the online learning approach~\cite{gu2021real, chen2022asymptotically}, we train on the $n$-th segment and evaluate on the \textbf{clicks of the $(n+1)$-th segment}, with metrics averaged across all segments.

We define four types of windows for \OursDataset: conversion attribution $\window^{attr}_\conversion$, refund attribution $\window^{attr}_\refund$, conversion observation $\window^{obs}_\conversion$, and refund observation $\window^{obs}_\conversion$. 
The conversion and refund attribution windows are both set to 3 days, defining the maximum delay for valid outcomes, while observation windows are 0.01 day (15 minutes) each. 

\subsection{Sample Delivery}\label{apx:sample-delivery}
Conversion events are fed into the model at the end of the conversion observation window if they occur within it; otherwise, they are added at the exact time of conversion. 
Refund events are fed into the model either at the time of actual refund or at the end of the refund observation window, whichever comes first. 
Notably, \textit{the refund observation window is a unique feature of \OursMethod}. 
Other methods only detect refunds at the actual refund time.
\subsection{Detailed Evaluation Metrics} \label{apx:metrics}
\label{sec:metrics}
We adopt the following metrics to evaluate both ranking and calibration performance in streaming NetCVR prediction:
\begin{itemize}[leftmargin=10pt,topsep=2pt]
  \item \textbf{AUC}: The area under the ROC curve, measuring the model's ability to distinguish between positive and negative samples in terms of pairwise ranking.

  \item \textbf{PRAUC}: The area under the precision-recall curve. Due to the highly imbalanced nature of streaming data, PRAUC is more sensitive than AUC to improvements on positive samples and better reflects practical utility.

  \item \textbf{NLL}: The negative log-likelihood, which evaluates the quality of predicted probabilities:
        \[
            \text{NLL} = -\frac{1}{N} \sum_{i=1}^N \left[ y_i \log \hat{p}_i + (1 - y_i) \log(1 - \hat{p}_i) \right],
        \]
    It penalizes miscalibration and is critical in cost-per-action (CPA) systems.

  \item \textbf{RI-AUC} (Relative Improvement in AUC): Following prior works~\cite{yang2021capturing, chen2022asymptotically}, we compute the relative improvement over the pre-trained model:
        \[
            \text{RI-AUC} = \frac{\text{AUC}_{\text{model}} - \text{AUC}_{\text{pre-trained}}}{\text{AUC}_{\text{oracle}} - \text{AUC}_{\text{pre-trained}}} \times 100\%,
        \]
    where $\text{AUC}_{\text{oracle}}$ is computed using ground-truth labels with full delay observation. RI-AUC measures how close a method approaches the oracle performance, with higher values indicating better recovery from delayed feedback.

  \item \textbf{PCOC} (Predicted Click-to-Conversion Over Calibration): We further evaluate calibration by computing the ratio of average predicted probability to empirical conversion rate:
        \[
            \text{PCOC} = \frac{\mathbb{E}[\hat{p}]}{\mathbb{E}[y]},
        \]
    A PCOC value close to 1 indicates well-calibrated predictions. Values $>1$ indicate overestimation (common in delayed feedback), while $<1$ indicate underestimation.
\end{itemize}

\subsection{Detailed Competitors}\label{apx:baselines}
The compared methods include:
\begin{itemize}[leftmargin=10pt,topsep=2pt]
  \item \textbf{Pre-trained}: The initial model trained on the pre-training set without any online updates. Serves as a static baseline to measure the value of streaming adaptation.

  \item \textbf{Oracle}: A privileged model fine-tuned with ground-truth labels after full observation (i.e., after $W_c + W_r$), representing the performance upper bound under complete feedback.

  \item \textbf{BDL(Batch Deep Learning)}: A daily batch model that waits for all feedback within attribution window, thus receiving delay-free labels. Simulates ideal offline training with no label censorship.

  \item \textbf{FNC}~\cite{ktena2019addressing}: Corrects selection bias by directly re-injecting delayed feedback samples into the training stream upon their arrival.

  \item \textbf{FNW}~\cite{ktena2019addressing}: Uses importance weighting to up-weight fake negatives and down-weight real negatives based on delay distribution.

  \item \textbf{DEFER}~\cite{gu2021real}: Replicates both delayed positive samples and confirmed real negative samples to mitigate feature distribution shift caused by asymmetric sample duplication.

  \item \textbf{DEFUSE}~\cite{chen2022asymptotically}: Introduces a four-way categorization of samples: Immediate Positive (IP), Fake Negative (FN), Real Negative (RN), and Delayed Positive (DP), with tailored modeling and reweighting for each type.

  \item \textbf{ES-DFM}~\cite{yang2021capturing}: The first method to introduce an explicit conversion observation window and perform debiasing within it using importance weighting. Models the trade-off between label completeness and data freshness via adaptive sample reweighting.

  \item \textbf{DDFM}~\cite{dai2023dually}: Proposes a hybrid modeling strategy: applies unbiased estimation for samples inside the observation window, and debiased modeling for those outside—without relying on survival models.

  \item \textbf{DSFN}~\cite{liu2023online}: Leverages transfer learning to capture shared representations across different types of streaming user feedback.

  \item \textbf{MISS}~\cite{liu2024online}: Employs multiple models trained on different observation windows and fuses their predictions via ensemble voting to improve robustness.
\end{itemize}

\subsection{Implement Details} \label{apx:implement}
We reproduce all competitors based on their official implementations. 
Since cascaded modeling outperforms direct modeling (\S~\ref{sec:analyze_divide}), all methods adopt a unified dual-tower architecture that independently estimates CVR and RFR. 
All models are initialized from the same pre-trained checkpoint and fine-tuned on identical streaming data, ensuring controlled comparisons under cascaded feedback. We implement \OursMethod and all baselines in PyTorch. 
Feature inputs are embedded and fed into fully connected networks with hidden sizes \{256, 256, 128\}, where each layer is followed by BatchNorm~\cite{2015Batch} and LeakyReLU~\cite{Maas0Rectifier}. 
All methods are optimized with Adam~\cite{2017Decoupled}, and hyperparameters are carefully tuned to report the best performance.

\subsection{Robustness to Refund Attribution Windows}\label{apx:refund-window}
To evaluate robustness under different refund attribution assumptions, we further test \OursMethod with varying refund attribution windows $\window^{attr}_\refund \in \{3\text{d}, 5\text{d}, 7\text{d}\}$ while keeping all other settings fixed.
This design intentionally introduces variation in the time allowed for attributing refunds, thereby simulating the practical uncertainty in real-world refund timing. It allows us to examine whether the performance is sensitive to such shifts in the attribution window.

\begin{figure}[t]
\centering
\includegraphics[width=0.8\columnwidth]{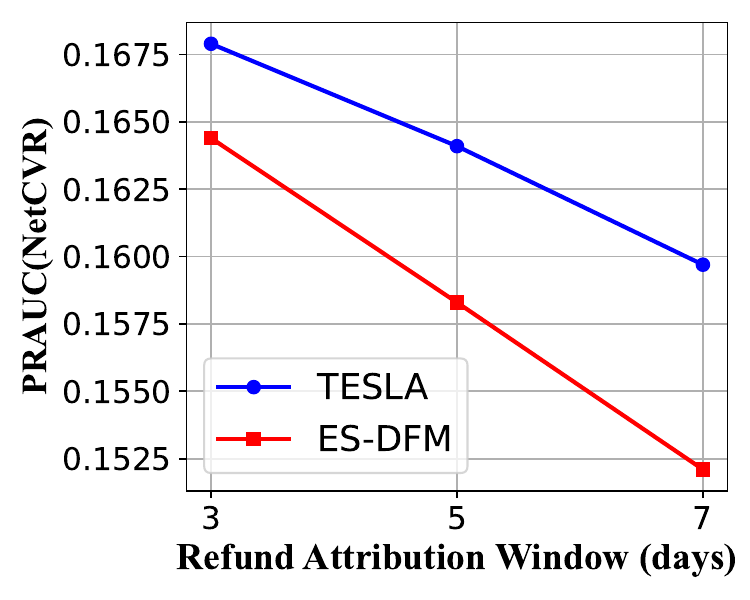}
\caption{Robustness study on refund attribution window. 
}
\label{fig:attr_window_ablation}
\end{figure}

The results indicate that \OursMethod consistently outperforms strong delayed-feedback baselines across all experimental settings.
As illustrated in Fig.~\ref{fig:attr_window_ablation}, the proposed \OursMethod consistently outperforms ES-DFM across different refund attribution windows in terms of NetCVR PR‑AUC, with relative improvements of approximately 2.1\%, 3.7\%, and 5.0\% for the 3‑day, 5‑day, and 7‑day windows, respectively.
Importantly, the performance gap remains stable as the refund window increases, suggesting that \textbf{\OursMethod is robust to variations in refund timing} and does not depend on finely tuned attribution assumptions.
These findings highlight  \textbf{\OursMethod’s effectiveness in handling realistic refund delay shifts}.

\end{document}